\documentclass[letterpaper]{article} 
\usepackage{aaai24}  
\usepackage{times}  
\usepackage{helvet}  
\usepackage{courier}  
\usepackage[hyphens]{url}  
\usepackage{graphicx} 
\usepackage{natbib}  
\usepackage{caption} 
\usepackage{algorithm}
\usepackage{algorithmic}
\def\ie{\textit{i.e.}} 
\def\etal{\textit{et~al.}} 
\def\eg{\textit{e.g.}}
\usepackage{newfloat}
\usepackage{listings}
\DeclareCaptionStyle{ruled}{labelfont=normalfont,labelsep=colon,strut=off} 
\floatstyle{ruled}
\newfloat{listing}{tb}{lst}{}
\floatname{listing}{Listing}
\usepackage{amsmath}
\usepackage{amssymb}
\def\ie{\textit{i.e.}} 
\def\etal{\textit{et~al.}} 
\def\eg{\textit{e.g.}}

\usepackage{booktabs}       
\usepackage{amsfonts}       
\usepackage{nicefrac}       
\usepackage{microtype}      
\usepackage{subcaption}
\usepackage{multirow}
\title{
Controllable 3D Face Generation with Conditional Style Code Diffusion
}
\author{
    Xiaolong Shen\textsuperscript{\rm 1,2}\footnote{Xiaolong Shen worked on this at his Alibaba internship.},
    Jianxin Ma\textsuperscript{\rm 2},
    Chang Zhou\textsuperscript{\rm 2},
    Zongxin Yang\textsuperscript{\rm 1}\footnote{Zongxin Yang is the corresponding author.}
}
\affiliations{
    \textsuperscript{\rm 1} ReLER, CCAI, Zhejiang University, China \\
    \textsuperscript{\rm 2} Alibaba Group, China \\
    \{sxlongcs, zongxinyang\}@zju.edu.cn, 
    \{majx13fromthu,ericzhou.zc\}@alibaba-inc.com
}

\begin{document}

\maketitle


\begin{abstract}
Generating photorealistic 3D faces from given conditions is a challenging task. Existing methods often rely on time-consuming one-by-one optimization approaches, which are not efficient for modeling the same distribution content, \eg, faces. 
Additionally, an ideal controllable 3D face generation model should consider both facial attributes and expressions.
Thus we propose a novel approach called TEx-Face~(\textbf{TE}xt \& \textbf{Ex}pression-to-Face) that addresses these challenges by dividing the task into three components, \ie, 3D GAN Inversion, Conditional Style Code Diffusion, and 3D Face Decoding. 
For 3D GAN inversion, we introduce two methods which aim to enhance the representation of style codes and alleviate 3D inconsistencies. 
Furthermore, we design a style code denoiser to incorporate multiple conditions into the style code and propose a data augmentation strategy to address the issue of insufficient paired visual-language data. 
Extensive experiments conducted on FFHQ, CelebA-HQ, and CelebA-Dialog demonstrate the promising performance of our TEx-Face in achieving the efficient and controllable generation of photorealistic 3D faces. The code will be available at \url{https://github.com/sxl142/TEx-Face}.
\end{abstract}

\begin{figure}[!t]
		\centering
		\includegraphics[width=0.47\textwidth]{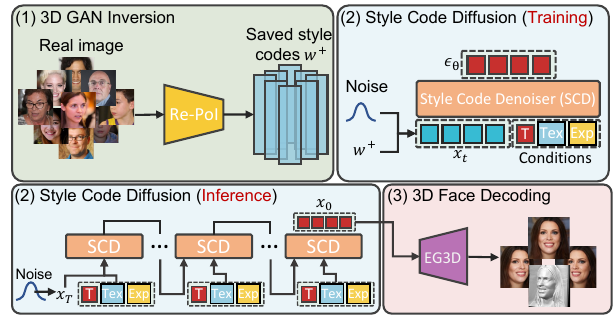}
		\caption{
            An overview of our pipeline. We train an inversion model called Re-PoI and save the style codes it infers. These saved codes are then used to train a style code denoiser with three conditions, \ie, time steps, text prompts, and expression codes. When inference, we decode the generated style codes into 3D faces using EG3D.
		}
		\label{fig:overview}
\end{figure}

 \begin{figure*}[!t]
		\centering
		\includegraphics[width=1.0\textwidth]{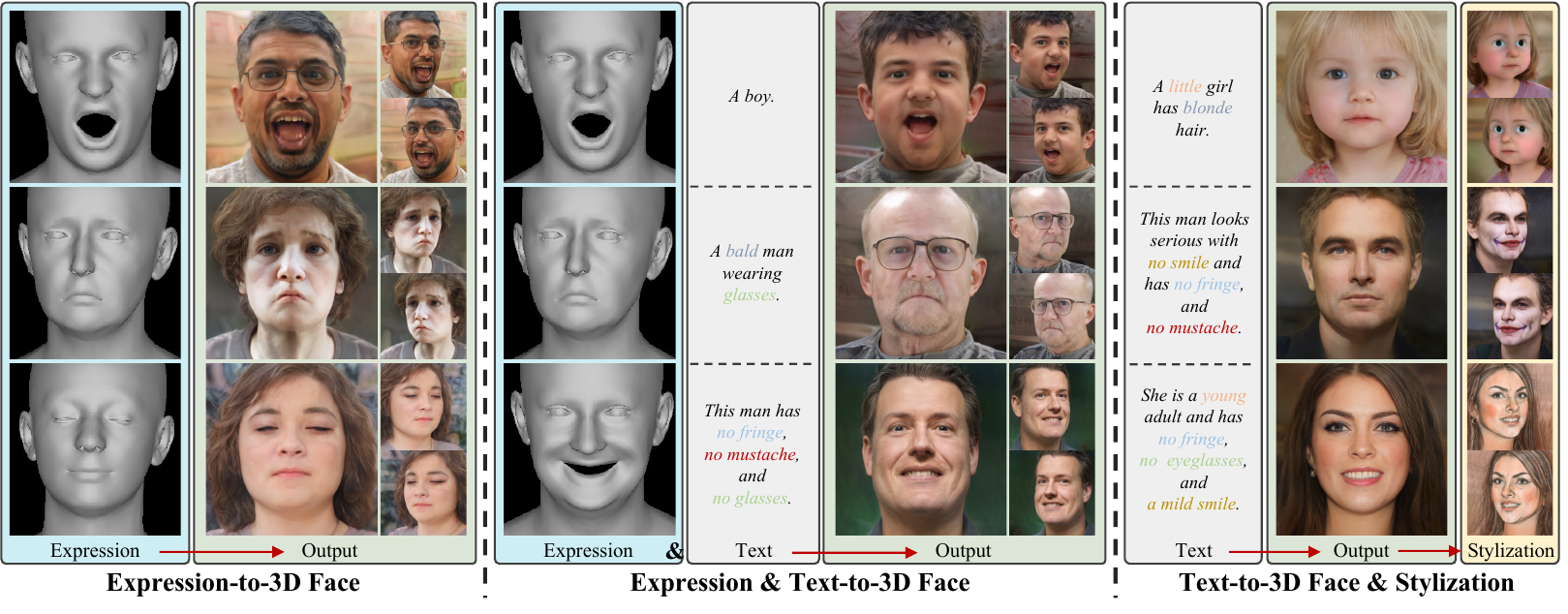}
		\caption{
            Our pipeline enables conditional 3D face generation using text, expression codes, and both of them. The stylization is achieved by StyleGAN-NADA~\cite{gal2021stylegannada}, a GAN-based style transfer method.
		}
		\label{fig:first_f}
\end{figure*}
\section{Introduction}
Face generation, as a longstanding task in image synthesis, is extensively utilized in artistic creation and portrait generation. Benefits from well-designed 2D/3D unconditional generation frameworks~\cite{Karras2020stylegan2,chan2022efficient}, current face generation models have achieved remarkable photorealism in both 2D and 3D. However, introducing multiple conditional representations~\cite{mkr} into face generation, particularly in the context of 3D, remains largely unexplored.

Existing Text-to-3D generation techniques~\cite{poole2022dreamfusion,huang2023avatarfusion, chen2023fantasia3d,xu2023seeavatar} typically involve training an initialized NeRF~\cite{mildenhall2020nerf} model to generate a specific object using pre-trained text-to-image diffusion models~\cite{latentdiff}. 
Although these approaches offer flexibility and can produce various objects, \textbf{generating a large number of objects from the same distribution is time-consuming}, such as generating 3D faces with one-by-one optimization schema. 
Besides, some text-to-3D face generation techniques~\cite{zhang2023dreamface, wu2023highfidelity} primarily focus on generating texture for parametric face models, \eg, 3D Morphable face model~\cite{3dmm}. While parametric models make it possible to explicitly control the expression, pose, and lighting of faces, the generated results often \textbf{lack realism} since the parametric geometry lacks finer details such as eyes, teeth, and hair.

Human faces consist of two main components, \ie, appearance and skeleton. The appearance represents the visual characteristics that differentiate individuals, while the skeleton is responsible for different poses and expressions. 
From this perspective, an ideal controllable generative model should be able to understand these inherent facial properties. 
Thus, we need to find a reasonable way to represent these properties for the generative model. 
The emergence of CLIP~\cite{clip} has bridged the gap between images and text, enabling successful text-conditioned generation like~\cite{nichol2022glide, ramesh2022hierarchical, saharia2022photorealistic}. This motivates us to choose the text to represent facial appearances.
Regarding the other component skeleton, a typical representation is a set of keypoints. Alternatively, we can leverage the parameters of the 3D Morphable face model, \ie, expression code. Compared to keypoints, these parameters can yield 3D meshes through the parametric model, offering greater flexibility in controlling the generated faces.

\begin{figure}[!t]
		\centering
		\includegraphics[width=0.45\textwidth]{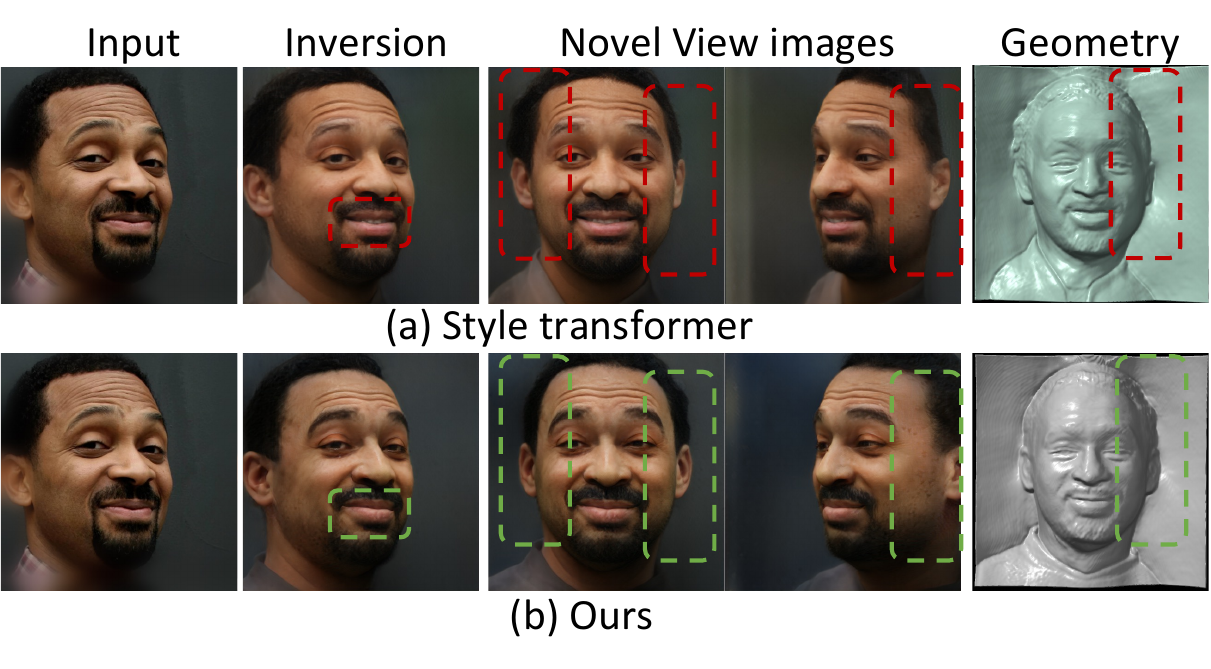}
		\caption{
            Simply extending the 2D Inversion method~\cite{hu2022style} leads to bad novel view synthesis, especially when inputting an image of side views. 
		}
		\label{fig:2dbadcase}
\end{figure}

Based on the aforementioned observations, we aim to explore an approach to address two key aspects: \textbf{1) generating photorealistic 3D faces efficiently}, and \textbf{2) controllable generation based on the given conditions}, \ie, text prompts and expression codes.
First, we consider that the current unconditional 3D GAN model~\cite{chan2022efficient} is a suitable foundation model, as it has demonstrated the ability to produce convincing 3D faces efficiently. 
Subsequently, choosing this model requires considering how to equip the 3D GAN with text-conditioned and expression-conditioned abilities. 
Upon analyzing the 3D GAN framework, we observe that the produced results are primarily influenced by style codes. 
Therefore, we resort to GAN Inversion technology~\cite{tov2021designing,alaluf2021restyle} to obtain the necessary style codes. 
Since there is a lack of encoder-based 3D GAN Inversion, we need to extend 2D inversion methods to the 3D domain. However, simply extending the 2D methods can lead to overfitting to the given camera pose, yielding undesirable novel view rendering, as shown in Figure~\ref{fig:2dbadcase}. 
Second, we need to consider how to inject multi-conditions into the obtained style codes. With the recent surge in diffusion-based generative models~\cite{ddpm,song2022denoising,zhou2023pyramid} applied to images, there is great interest in exploring the potential of diffusion models for learning style codes, given their ability to facilitate flexible condition injections. However, aligning the given conditions with semantic-agnostic style codes remains a challenging open problem.

To address these issues, we propose a novel framework named TEx-Face, as shown in Figure~\ref{fig:overview}, involving three components. \textbf{1) 3D GAN Inversion.} We design two methods for 3D GAN Inversion, \ie, Pose-guided Inversion~(PoI) Pretraining and Refined Pose-guided Inversion~(Re-PoI) Funetuning, targeting to alleviate the 3D inconsistency and enhance the representation of style codes. 
In pretraining, we expect that the images' style codes are consistent under the different camera views for one identity. 
Thus, we need to project each view's style codes onto a canonical style code.
Here, we choose style codes under the front view as canonical codes due to the largest visible facial appearance. 
Specifically, due to the lack of multi-view images, we first generate synthetic canonical style codes using the Mapping Network in 3D GAN and then use these codes to render images under randomly sampled camera poses. After that, we can leverage the PoI to learn a mapping that projects the different views' style codes onto one canonical style code, yielding a view-invariant code. 
In the finetuning stage, we apply coarse-to-fine schema, incorporating a new branch to learn the difference between the reconstructed and input images and complement details for high-quality reconstruction. 
\textbf{2) Style Code Diffusion.} We introduce a style codes denoiser to decoupled inject the text embedding extracted by CLIP~\cite{clip} and expression code extracted by EMOCA~\cite{EMOCA} into the style codes for learning the distribution alignment. 
Moreover, considering the insufficient of paired visual-language face data, we propose a data augmentation strategy for improving the diversity of the origin dataset, which utilizes a manipulation method and an image captioning method to automatically generate paired data. 
\textbf{3) 3D Face Decoding.} The style codes are crucial for this task, they directly affect the quality of the generated results. Hence, we select the most advanced method, EG3D~\cite{chan2022efficient}, as our decoding model. This method is based on the architecture of StyleGAN2~\cite{Karras2020stylegan2}, which results in a well-behaved style code space.

We conduct extensive experiments on several datasets, \ie, FFHQ~\cite{abdal2019image2stylegan}, CelebA-HQ~\cite{karras2018progressive}, and CelebA-Dialog~\cite{jiang2021talk}. Experiments show that our method improves the image quality and alleviates the 3D inconsistency for 3D GAN Inversion, as shown in Figure~\ref{fig:2dbadcase}. 
To evaluate the 3D Face generation quality, we provide various metrics, including FID, CLIP score, expression score, and user studies. The results indicate that our model is capable of producing photorealistic 3D faces that align with the given text prompts and expression codes.

Overall, our contributions are summarized as follows,
\begin{itemize}
    \item To our best knowledge, we make the first attempt to explore controllable 3D face generation using multi-conditions and one-time optimization method, which decomposes it into three parts, \ie, 3D GAN Inversion, Style Code Diffusion, and 3D Face Decoding. 
    \item We design two methods for 3D GAN Inversion, \ie, Pose-guided Inversion Pretraining and Refined Pose-guided Inversion Funetuning, to enhance the style codes and alleviate the 3D appearance inconsistency. 
    \item We introduce a style code denoiser to inject text embedding and expression code into the style codes. Moreover, we propose a data augmentation strategy to generate paired data automatically.
    \item Extensive experiments show our framework achieves promising conditional 3D face generation results.
\end{itemize}

\section{Related Work}
\subsection{2D\&3D GAN Inversion}
GAN Inversion~\cite{zhu2016generative,tov2021designing,roich2022pivotal,wang2021HFGI,hu2022style,xie2022highfidelity,yin20233d} focuses on finding a semantically rich latent space for the images, which can be further used to edit the images via some latent space manipulation methods~\cite{wei2022hairClip,patashnik2021styleClip,zhu2022one, shen2020interpreting,zhu2020domain}. 2D GAN Inversion method are mainly based on StyleGAN2~\cite{Karras2020stylegan2} owing to the well-designed architecture.
There are three types of Inversion methods, \ie, optimize-based~\cite{xie2022highfidelity,yin20233d, wu2021stylespace}, encoder-based~\cite{tov2021designing, wang2021HFGI,hu2022style,alaluf2021restyle}, and hybrid-based methods~\cite{alaluf2021restyle,chai2021ensembling,ko20233d}. Optimize-based methods directly optimize a latent code for a specific image, which makes the reconstructed images more realistic and fidelity but usually takes a long time.
Encoder-based methods only spend time on training, while the quality of reconstructed images is worse than optimize-based methods. Hybrid-based methods leverage the latent codes inferred by encoder-based methods as initial latent codes for optimize-based methods. However, these method is still time-consuming. 
Recently, EG3D~\cite{chan2022efficient} leverages StyleGAN2 generator as a backbone to yield tri-plane for 3D face generation, which makes it possible that can simply extend 2D GAN Inversion methods by changing the StyleGAN2 with EG3D due to inherent similarity. However, simply extending 2D methods results in 3D inconsistency. While some works~\cite{yin20233d,xie2022highfidelity,ko20223d} target to solve this problem, these methods are optimization-based, which is not suitable for generating a large number of style codes. 

\begin{figure*}[!t]
		\centering
		\includegraphics[width=1.\textwidth]{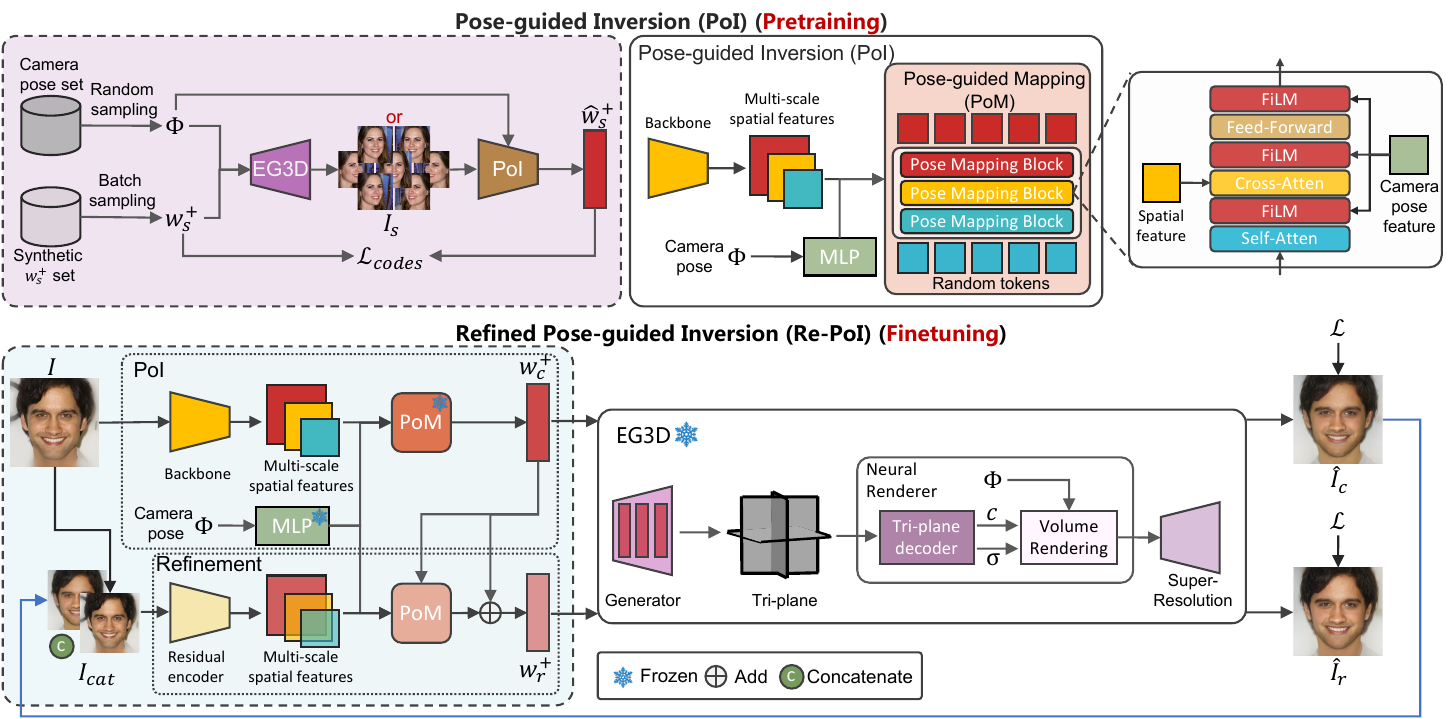}
		\caption{
  An overview of our inversion method. In pretraining, we first leverage the synthetic multi-view images to learn a mapping that projects the style codes under different views onto one style code, which yields a view-invariant style code, thereby alleviating 3D inconsistency. 
  In finetuning, we freeze the learned mapping~(PoM) and MLP in the PoI and further append the refinement branch to improve the quality of the style codes. Note the PoM of the refinement branch is used for training.
    }
		\label{fig:inversion_pipline}
\end{figure*}

\subsection{Conditional Generation}
\noindent\textbf{2D generation.}
Earlier works~\cite{xu2018attngan,yang2019very} are mainly constructed on Generative Adversarial Networks (GANs)~\cite{goodfellow2014generative}. The recent surge in Diffusion models~\cite{ddpm, song2022denoising,zhou2023pyramid} is a class of generative models that excel at generating realistic images through iterative refinement of a sampled noise. DDPM~\cite{ddpm} employs a dual Markov chain approach for unconditional image generation.
Imagen~\cite{saharia2022photorealistic} explores large language models~\cite{raffel2020exploring} trained on text data as text encoders for text-to-image diffusion models. Unlike above, stable-diffusion~\cite{latentdiff} applies the diffusion paradigm to the latent space of a Variational autoencoder (VAE) and introduces cross-attention layers into UNet for additional conditions.

\noindent\textbf{3D generation.} 
In recent years, the field of text-to-image generative modeling~\cite{nichol2022glide,wang2022high,latentdiff} has achieved significant advancements, leading to a surge of interest in text-to-3D generation~\cite{poole2022dreamfusion,huang2023avatarfusion,xu2023seeavatar}.
CLIP-Forge~\cite{sanghi2022clip} capitalizes on the shared text-image embedding space of CLIP to train on image embeddings and utilize text embeddings for inference, thereby enabling text-to-3D capabilities. 
DreamFusion~\cite{poole2022dreamfusion} employs a pretrained text-to-image diffusion model as a potent image prior and introduces a Score-Distillation loss to supervise the generation of 3D objects represented by Neural Radiance Fields (NeRFs~\cite{mildenhall2020nerf}). 
Following DreamFusion, AvartarFusion~\cite{huang2023avatarfusion} and SEEAvatar~\cite{xu2023seeavatar} focus on high-quality 3D human generation by combining parametric human models and text-to-image diffusion models.
For text-to-3D face generation, current methods~\cite{zhang2023dreamface, wu2023highfidelity} mainly build on a 3D Morphable face model and generate 3D faces with geometry and texture. Owing to the parametric model, these models can explicitly control the expression, pose, and so on. However, the generation results usually lack details, \eg, hair, and look unnatural.

\section{Method}
In this section, we first recap the EG3D~\cite{chan2022efficient} and diffusion models in preliminary, and then we introduce the designed inversion model and style code diffusion model.
\subsection{Preliminary}
\noindent\textbf{EG3D.} As shown in Figure \ref{fig:inversion_pipline}, EG3D is composed of a Mapping Network and a Synthesis Network. The Mapping Network generates a style code $w^+, w^+ \in \mathbf{R}^{14 \times 512}$ by feeding a combination of a sampled Gaussian vector $z, z\in \mathbf{R}^{512}$ and a camera pose $ \phi, \phi \in \mathbf{R}^{25}$ sampled from datasets. 
Here we only draw the Synthesis Network because the inversion process replaces the Mapping Network to yield style codes. 
After that, the Synthesis Network produces a Tri-plane representation via a series of convolution layers modulated by the style code $w^+$. 
The resulting Tri-plane representation is then decoded by the Neural Renderer, producing coarse images. Finally, the generated coarse images are refined by a Super-Resolution Network, enhancing their quality.

\noindent\textbf{Diffusion models.} Recently, diffusion models~\cite{ddpm,ho2022classifierfree,song2022denoising} have demonstrated remarkable capabilities in conditional or unconditional generation tasks. 
Diffusion models are a class of probabilistic models that involve two stages, \ie forward and reverse processes. The forward process works by gradually adding noise to the data in $T$ steps, generating a sequence of noise samples $x_1, ..., x_T$. It follows the rule of the Markov chain. The resulting noise sample $x_t$ can be defined as follows,
\begin{small}
    \begin{align}
    q(x_t|x_{t-1}) = \mathcal{N}(x_t; \sqrt{1-\beta_t} x_{t-1}, \beta_t \mathbf{I})  \\
    x_t = \sqrt{\alpha_t} x_{0} + \sqrt{1 - \alpha_t} \epsilon, \space \space \space \epsilon \sim \mathcal{N} (\mathbf{0}, \mathbf{I}), 
    \end{align}
\end{small}
where $\alpha_t = \prod_{i=1}^t (1 - \beta_i)$, $\{\beta_t\}_{t=1}^T$ are fixed or learned variance schedule. In the reverse stage, we need to learn a model $p_\theta(x_{t-1} | x_t)$ to approximate the conditional probabilities $q(x_{t-1} | x_t)$ because we cannot easily estimate the distribution of the whole dataset. Ho~\etal find the training objective can be simplified to make the diffusion model learn better. The function is as follows, 
\begin{small}
    \begin{equation}
    \begin{aligned}
     \min_{\theta}\mathbb{E}_{x_0 \sim q(x_0), \epsilon \sim \mathcal{N}(\mathbf{0,I}), t\sim [1, T]} || \epsilon -\epsilon_\theta(x_t, t) ||^2_2.    
    \end{aligned}
\end{equation}
\end{small}

\subsection{3D GAN Inversion}
\noindent\textbf{Pose-guided Inversion~(PoI) Pretraining.}
We demonstrate that simply utilizing the 2D inversion method can result in overfitting to the input view as shown in Figure~\ref{fig:2dbadcase}, leading to 3D inconsistency when rendering images from novel views. 
To address this, we propose an explicit learning method that aims to project style codes under different views onto a single style code, producing a view-invariant style code. We consider a style code under the front view as canonical code.
However, we lack multi-view images of one identity to model this mapping. To achieve this, we leverage the pretrained EG3D to produce synthetic multi-view images by using this canonical style code as depicted in Figure~\ref{fig:inversion_pipline}.

Specifically, we first sample 50K style codes $w_s^+$ from the Mapping Network in EG3D and save them. Subsequently, we batch sample $w_s^+$ which are then used to render images with randomly sampled camera poses~$\Phi$.
Lastly, the produced images will be encoded as style codes~$\hat{w}_s^+$ through our proposed PoI. The PoI consists of a backbone and a Pose-guided Mapping Network~(PoM). Given a synthetic image $I_s$, the backbone extracts multi-scale features~$M$ from this image, and then the features with the camera feature will be fed into the PoM. The PoM has two important layers, \ie, a cross-attention layer and a Feature-wise Linear Modulation~(FiLM) layer. The cross-attention layer is used for decoding multi-scale features, while the FiLM is responsible for mapping these decoded features to the expected canonical space by conditioning the camera feature.
The main processes are as follows,
\begin{small}
    \begin{equation}
    \begin{aligned}
        CrossAtten(Q^c, K_{M^c_i}, V_{M^c_i})&
			= Softmax (\frac{Q^c K_{M^c_i}^T}{\sqrt[]{C}})V_{M^c_i}, \\
   FiLM(f_p,f_c)& = \gamma(f_c)f_p + \beta(f_c),
    \end{aligned}
\end{equation}
\end{small}
where $M^c_i$ is an $i_{th}$-scale spatial feature, $\gamma$ and $\beta$ are Linear function. The camera feature $f_c$ is extracted by the Multilayer Perceptron~(MLP). The $f_p$ is the feature of the previous layer. Finally, The PoM transforms the random tokens to the coarse style codes $w_c^+$ by several Pose Mapping Blocks. 

\noindent\textbf{Refined Pose-guided Inversion~(PoI) Finetuning.} 
In pretraining, we train the PoI on synthetic data. Thus we need to finetune the PoI on real data. However, fully finetuning the model will discard the learned projection ability that maps the style codes of multi-view images to the canonical style code. Therefore, we freeze the PoM and only finetune the backbone. Besides, simply finetuning merely yields coarse codes, lacking some details. Hence, we propose to complement details by using a coarse-to-fine schema.
Similar to the PoI, the refined module involves a PoM and residual encoder.

Specifically, we concatenate the coarse image $\hat{I}_c$ and ground truth $I$ as the input $I_{cat}$ of the Residual Encoder~(RE). Similarly, the encoder extracts multi-scale features $M_r$ for the Re-PoM. In the first block of the Re-PoM, it will use the coarse style codes as input to query the extracted residual multi-scale features. Finally, the output of Re-PoM will be added with the coarse style codes as the final results $w_r^+$.
The main processes are formulated as follows.
\begin{small}
    \begin{equation}
    \begin{aligned}
         &M_r = RE(concat(I, \hat{I}_c)), \\
      CrossAtten&(Q_r, K_{M^r_i}, V_{M^r_i})
        = Softmax (\frac{Q_r K_{M_i^r}^T}{\sqrt[]{C}})V_{M_i^r}, \\
     &w_r^+ = w_c^+ + RePoM(M_r, w_c^+).
    \end{aligned}
\end{equation}
\end{small}

\noindent\textbf{Training objective.} 
To learn a view-invariant style code in the pretraining stage, we apply loss on the style codes.
\begin{small}
    \begin{align}
     \mathcal{L}_{codes} = || w_{s}^+ -  \hat{w}_{s}^+ ||_2
\end{align}
\end{small}
In finetuning, we apply MSE, LPIPS, and ID loss for reconstructed images. Besides, inspired by ~\cite{abdal20233davatargan}, we apply the depth loss on the reconstructed images, which will further improve the quality of geometry. 
\begin{small}
    \begin{equation}
    \mathcal{L} = \lambda_{rec} * \mathcal{L}_{rec} + \lambda_{lpips} * \mathcal{L}_{lpips} +
    \lambda_{id} * \mathcal{L}_{id} + \lambda_{dep} * \mathcal{L}_{dep}
\end{equation}
\end{small}

\begin{figure}[!t]
    \centering
    \includegraphics[width=0.42\textwidth]{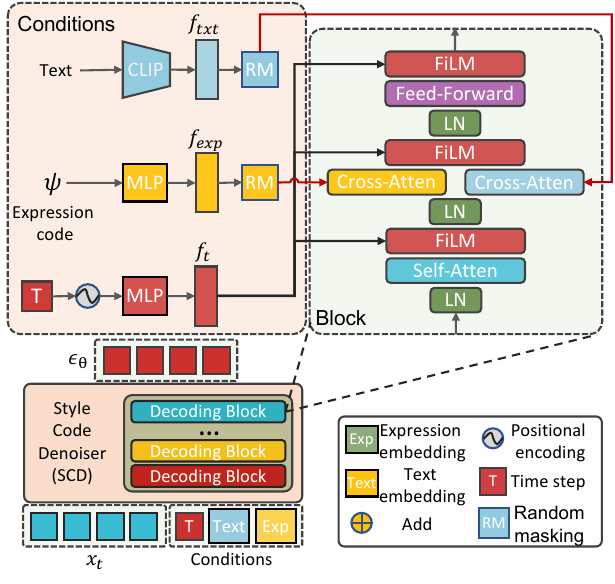}
    
    \caption{ A schematic of Style Code Denoiser. }
    \label{fig:scd}
\end{figure}

\subsection{Conditional Style Code Diffusion}
Considering the facial ingredients, \ie, appearance and expression, we expect the model can be controlled by these two aspects. Thus we propose incorporating text embedding and expression codes to guide the generation of appearance and skeleton, respectively. 
However, incorporating these conditions into the model flexibly and effectively poses a challenge. Upon analyzing the overall structure of the EG3D, we identify the style code inferred by the Mapping Network of the EG3D as a crucial component in determining the generated results. 
Thus we design a style code denoiser to inject these conditions into this style code.

\noindent\textbf{Style Code Denoiser.}
To achieve a multi-conditioned model, we structurally decouple the conditions by employing a separated cross-attention layer. 
Concretely, as shown in Figure \ref{fig:scd}, there are three conditions, \ie, text, expression code, and time step.
To model the text embedding~$f_{txt}$, we utilize the CLIP text encoder. CLIP enables us to encode textual information into a meaningful representation that can guide the generation of facial appearance.
For the expression embedding, we first employ EMOCA~\cite{EMOCA} to predict a set of expression code, denoted as $\psi, \psi \in \mathbf{R}^{50}$ and then utilize an MLP to obtain the final representation$f_{exp}$. 
To incorporate the time step information, we utilize positional encoding and an MLP, which allows us to represent the temporal aspect of the generation process.
Next, we apply random masking to both text and expression embeddings~($f_{txt}, f_{exp}$) to implement classifier-free guidance. Then these features are used as keys and values for extracting information through a separated cross-attention layer in the decode block. 
The time embedding is responsible for transforming the results to the given time steps by using FiLM layers.
Since we need to inject two conditions into the model, the training objective will be changed with multi-conditioned as follows.
\begin{small}
    \begin{equation}
    \begin{aligned}
     \min_{\theta}\mathbb{E}_{x_0 \sim q(x_0), \epsilon \sim \mathcal{N}(\mathbf{0,I}), t\sim [1, T]} || \epsilon -\epsilon_\theta(x_t, t, c_{txt}, c_{exp}) ||^2_2.    
    \end{aligned}
\end{equation}
\end{small}

\noindent\textbf{Data augmentation.}
Existing visual-language face dataset Celeba-Dialog~\cite{jiang2021talk} is built on the CelebaA-HQ~\cite{karras2018progressive}, which lacks some attribute annotation, \eg, hair color or hairstyle, and mainly consists of adults, lacking children and old people. Based on these observations, we try to bridge these gaps by using existing manipulation methods and image captioning methods to automatically generate paired data. Specifically, we implement HairCLIP~\cite{wei2022hairClip} on the CelebA-HQ training set to obtain 2D text-paired images of different hair colors and hairstyles. Then we leverage our method to inverse these images to obtain style codes. Moreover, we use captioning model~\cite{li2023blip2} to generate text for FFHQ~\cite{abdal2019image2stylegan}. The benefits of using captioning models are that we can use more face data, thus the trained model can handle a wider face distribution and improve facial diversity. For text-guided manipulation methods, we can use them to manipulate face images to obtain more facial attributes paired with text. 

\section{Implementation Details}
We use Adam~\cite{kingma2017adam} optimizer with linear warm-up and Cosine Annealing~\cite{loshchilov2017sgdr} scheduler.
We set the loss weights as follows: $\lambda_{rec} = 1, \lambda_{lpips} = 0.8, \lambda_{id} = 0.2$. We use four Nvidia Tesla V100~(16G) with batch size 8 to train our inversion model, and with batch size 256 for style code diffusion. For data processing, we follow EG3D, using off-the-shelf pose estimators to extract approximate camera extrinsic. 

\section{Experimental Results}
\begin{table}[!t]
  \centering
  \begin{tabular}{l|c|c|c|c}
    \toprule[2pt]
    Method & MSE $\downarrow$ & LPIPS $\downarrow$ & ID $\downarrow$ & FID $\downarrow$\\
    \midrule
    e4e & 0.0354 & 0.1970 & 0.1903 & 33.97 \\
    ST & 0.0259 & \textbf{0.1677} & 0.2199 & 34.59\\
    Ours & \textbf{0.0252} & 0.1721 & \textbf{0.1788} & \textbf{30.15} \\
    \bottomrule[1pt]
  \end{tabular}%
    \caption{GAN Inversion. 
    }
    \label{tab:inversion}
\end{table}

\begin{table}[!t]
    \centering
    \small
    \begin{tabular}{l|cc|cc|cc}
        \toprule[2pt]
        \multirow{2}{*}{ } & \multicolumn{2}{c|}{MSE$\downarrow$}  & \multicolumn{2}{c|} {LPIPS$\downarrow$}  & \multicolumn{2}{c}{ID$\downarrow$} \\
        & Self & Others & Self & Others & Self & Others  \\
        \midrule
        
       ST & 0.014 & 0.168 &  0.107 & 0.477 & 0.175 & 0.375 \\
        Ours & \textbf{0.012} & \textbf{0.048} & \textbf{0.105} & \textbf{0.189} & \textbf{0.163} & \textbf{0.244} \\
        
        \bottomrule[1pt]
    \end{tabular}%
    \caption{
        Evaluation of 3D consistency.
    }
    \label{tab:mead}
\end{table}

\begin{figure*}[!t]
		\centering
		\includegraphics[width=0.9\textwidth]{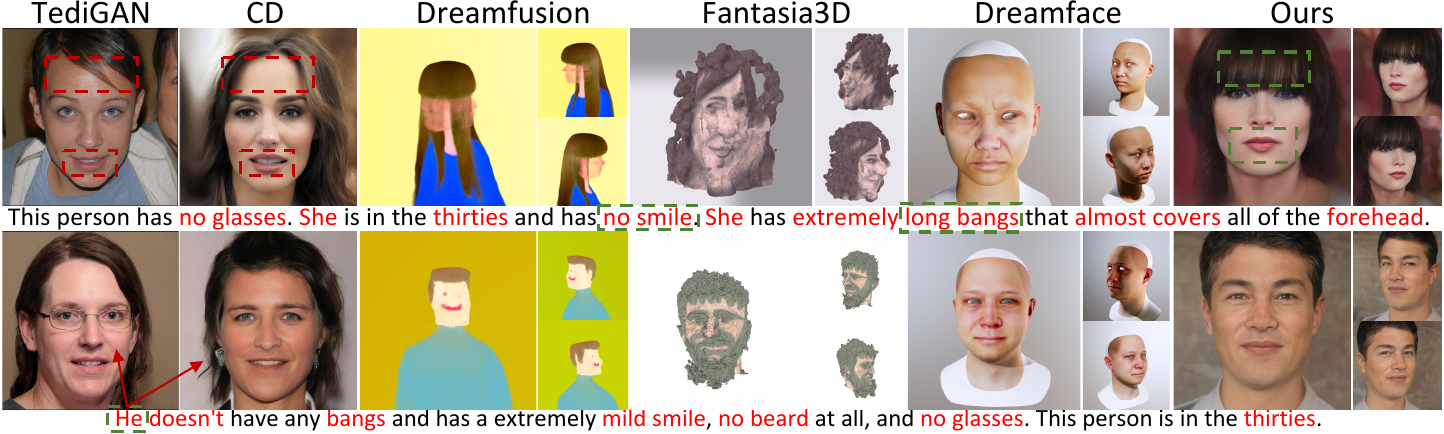}
		\caption{
            Visual comparison of Text-to-Face.
		}
		\label{fig:vsp}
\end{figure*}

\noindent\textbf{Datasets \& Metrics.}
We train our inversion model on FFHQ~\cite{abdal2019image2stylegan} and test it on CelebaA-HQ~\cite{karras2018progressive} test set. We use CelebA-Dialog~\cite{jiang2021talk} and some data processed by our proposed data augmentation strategy to train our diffusion model.
We evaluate inversion methods by Mean Squared Error~(MSE), Perceptual Similarity~(LPIPS~\cite{zhang2018unreasonable}), FID~\cite{heusel2017gans}, and identity similarity~(ID). For generation quality, we use FID, CLIP score, expression score, and user studies. 
\begin{table}[!t]
  \centering
  \begin{tabular}{l|c|c|c}
    \toprule[2pt]
    Method & Type & FID $\downarrow$ & CLIP score $\uparrow$ \\
    \midrule
    TediGAN & 2D & 157.81 & 24.27 \\
    CD & 2D & 111.36 & 24.51 \\
    Ours & 3D & \textbf{52.91} & \textbf{24.83} \\
    \bottomrule
\end{tabular}
\caption{Conditional face generation. 
}
\label{tab:condition_gen}
\end{table}
\begin{table}[!t]
  \centering
  \begin{tabular}{l|c|c}
    \toprule[2pt]
     & Text & Expression  \\
    \midrule
    Acc(\%)$\uparrow$ & 79.3 & 60.0 \\
    \bottomrule
\end{tabular}
\caption{User studies.}
\label{tab:user_st}
\end{table}
\begin{table}[!t]
    \centering
    \small
    \setlength{\tabcolsep}{3.5 pt}
    \begin{tabular}{l|c|c|c|c|c}
        \toprule[2pt]
         & Happy & Sad & Surprise & Disgust & Anger \\
        \midrule
        Confidence $\uparrow$ & 91.49 & 53.97 & 90.65 & 35.92 & 34.26\\
        \bottomrule
    \end{tabular}
    \caption{
        Expression score of our method.
    }
    \label{tab:expression_score}
\end{table}

\noindent\textbf{Setting.} Optimization-based GAN inversion methods~\cite{roich2022pivotal,xie2022highfidelity} are time-consuming when applied to obtain a large number of style codes. Thus we only focus on encoder-based inversion methods.
Due to the lack of released encoder-based 3D inversion methods, we re-implement 2D state-of-the-art inversion methods~\cite{tov2021designing,hu2022style} for 3D GAN inversion by simply replacing StyleGAN2 with EG3D.

\subsection{Quantitative Comparison.}
\noindent\textbf{GAN Inversion.} As shown in Table~\ref{tab:inversion}, Style Transformer~(ST)~\cite{hu2022style} obtains the lowest LPIPS metric but other metrics are worse than our method. 
Moreover, ST suffers from 3D inconsistency, as shown in Figure~\ref{fig:2dbadcase}. To evaluate the 3D consistency, we conducted experiments on the MEAD~\cite{wang2020mead} dataset in Table~\ref{tab:mead}. ``Self” measures self-reconstruction. ``Others” measures novel view synthesis, which means rendering images under different views with the input. MEAD has different view images of one identity. Thus, we can compute metrics with the ground truth.
It shows our model outperforms the ST, \textbf{particularly in ``others”}. It also demonstrates the 2D method is easier to overfit to the input view, leading to worse performance in novel view synthesis.

\noindent\textbf{Conditional 3D Face Generation.} 
Due to a lack of similar methods, we compare our method with 2D conditional generation methods~\cite{xia2021tedigan,huang2023collaborative} using the FID metric for generation quality and CLIP score for text-image matching. The results, presented in Table~\ref{tab:condition_gen}, demonstrate that our method achieves the best performance.
Furthermore, we evaluate our expression-to-3D faces by choosing five emotions~(Happy, Sad, Surprise, Disgust, and Anger) and use the recognition model of EMOCA~\cite{EMOCA} to test the generation results. 
The average confidence of the corresponding emotion class is provided in Table~\ref{tab:expression_score}. Our model achieves high confidence for Happy and Surprise expressions. While Sad, Disgust, and Anger do not reach high confidence, they are significantly better than a random probability $1 / 11 * 100 = 9 \% $ since the recognition model has 11 classes. Moreover, we consider the data imbalance also influences the performance.
As shown in Table~\ref{tab:user_st}, user studies are provided to validate text-image matching and expression-image matching. The results show that our model can generate plausible 3D faces based on given conditions.

\noindent\textbf{Inference time.} PTI~\cite{roich2022pivotal}, an optimization-based GAN inversion method, requires about 120s to process a single sample. Consequently, employing this method for a large number of samples is time-consuming. In contrast, our method takes only 0.10 seconds per sample.
Regarding style code diffusion, we employ a DDIM sampler with a sampling step of 100. This process takes approximately 5.1s.


\begin{table}[!t]
    \small
    \centering
    
    \setlength{\tabcolsep}{10 pt}
    \begin{tabular}{l|c|c}
        \toprule[2pt]
        Type & Expression score $\uparrow$ & CLIP score $\uparrow$ \\
        \midrule
        Coupled & 58.38 & 20.62 \\
        Decoupled & \textbf{59.70} & \textbf{21.86} \\
        \bottomrule
    \end{tabular}
    \caption{
        Different types of cross-attention layers. 
    }
    \label{tab:cross_atten}
\end{table}

\begin{figure}[!t]
		\centering
		\includegraphics[width=0.4\textwidth]{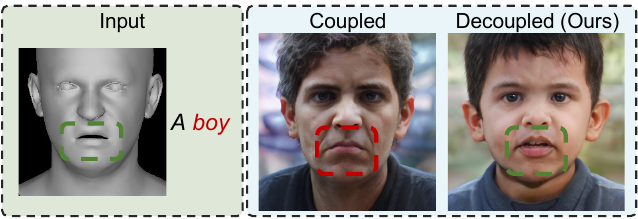}
		\caption{
            Visual comparison with Coupled and Decoupled cross-attention layers.
		}
		\label{fig:ab_cross}
\end{figure}

\subsection{Qualitative Comparison.}
\noindent\textbf{3D GAN inversion.} In Figure~\ref{fig:2dbadcase}, Our method shows improvements in facial symmetry and overall geometric accuracy compared with simply extending the 2D method to a 3D domain. More results are placed in the Supplementary. 

\noindent\textbf{Conditional 3D face generation.} 
As illustrated in Figure~\ref{fig:first_f}, we provide the results of text-to-3D face, expression-to-3D face, and text \& expression-to-3D face, which demonstrate that our pipeline is capable of generating plausible 3D faces. Moreover, we utilize StyleGAN-NADA~\cite{gal2021stylegannada}, a GAN-based style transfer method, which enables us to generate 3D faces in different domains by simply replacing the original EG3D model with a style-EG3D model trained using this method. 
For visual comparison with other methods, we show it in Figure~\ref{fig:vsp}, our pipeline surpasses other methods both in quality and text-face matching.

\subsection{Ablation Study.}

\noindent\textbf{3D GAN inversion.} We provide the results without refinement in the Re-POI finetuning stage in Table~\ref{tab:ab_inver}. It demonstrates that the refinement module significantly improves the performance. More results are in the supplementary.

\noindent\textbf{Conditional 3D face generation.}
We conduct experiments that apply different cross-attention layers as shown in Table~\ref{tab:cross_atten} and Figure~\ref{fig:ab_cross}. 
Coupled means using the same cross-attention to model text and expression embedding, while Decoupled represents using separated cross-attention.
The results show that the decoupled layer yields better performance and produces 3D faces that align well with the given conditions.

\begin{table}[!t]
    \small
    \centering
    \begin{tabular}{l|c|c|c|c}
        \toprule[2pt]
        Method & MSE $\downarrow$ & LPIPS $\downarrow$ & ID $\downarrow$ & FID $\downarrow$\\
        \midrule
        w/o refinement & 0.0404 &  0.2128 & 0.2156 & 35.90 \\
        w/ refinement & \textbf{0.0252} & \textbf{0.1721} & \textbf{0.1788} & \textbf{30.15} \\
        \bottomrule[1pt]
    \end{tabular}%
    \caption{
        Ablation studies of Re-PoI.
    }
    \label{tab:ab_inver}
\end{table}

\section{Conclusion}

We propose TEx-Face for conditional 3D face generation, consisting of three components, \ie, 3D GAN inversion, Conditional Style Code Diffusion, and 3D Face Decoding. 
Under the framework, we design two methods for 3D GAN inversion to enhance the representation of style codes and alleviate the 3D inconsistency. 
For the multi-conditioned generation, we propose a style code denoiser that successfully decoupled models three conditions, \ie, time step, text, and expression code. 
Moreover, we introduce a data augmentation strategy to generate paired data automatically. 
Based on these designs, our pipeline shows remarkable results.
  
\noindent\textbf{Limitations and Future Work.} The generation quality is highly dependent on the capability of EG3D and the inversion model. Moreover, the training data are imbalanced in expression. Thus, our model may focus on the majority of expressions, \eg, happy. 
The data augmentation strategy relies on manipulation and captioning methods. Hence, the ability of these models dominates the quality of the generated paired data. 
We will extend this framework to some methods~\cite{An_2023_CVPR, gao2022get3d} having a similar architecture with EG3D and will explore data augmentation strategy with our method in face detection (\eg, ~\cite{zhou2021face}).

\section*{Acknowledgements} 
This work was supported by the Fundamental Research Funds for the Central Universities~(No.~226-2023-00051).

\bibliography{aaai24}

\end{document}